\documentclass[10pt,twocolumn,letterpaper]{article}

\usepackage{cvpr}              

\usepackage[normalem]{ulem}
\newcommand{\methodname}{\textit{DetDiffusion}\xspace}
\newcommand{\uit}[1]{\textit{\uline{#1}}}

\definecolor{cvprblue}{rgb}{0.21,0.49,0.74}
\usepackage[pagebackref,breaklinks,colorlinks,citecolor=cvprblue]{hyperref}
\usepackage{xcolor} 
\usepackage{colortbl}

\definecolor{backcolor}{RGB}{232, 242, 255}
\definecolor{improvecolor}{RGB}{112, 173, 71}
\definecolor{baselinecolor}{gray}{.9}

\def\paperID{***} 
\def\confName{CVPR}
\def\confYear{2024}

\title{\methodname: Synergizing Generative and Perceptive Models for\\ Enhanced Data Generation and Perception}

\author{
Yibo Wang$^{1}\textsuperscript{*}$\quad
Ruiyuan Gao$^{2}\textsuperscript{*}$\quad
Kai Chen$^{3}\textsuperscript{*}$\quad
Kaiqiang Zhou$^{4}$\quad
Yingjie Cai$^{4}$\\
Lanqing Hong$^{4}$\quad
Zhenguo Li$^{4}$\quad
Lihui Jiang$^{4}\textsuperscript{\Letter}$\quad
Dit-Yan Yeung$^{3}$\quad
Qiang Xu$^{2}$\quad
Kai Zhang$^{1,5}\textsuperscript{\Letter}$\\
$^1$Tsinghua University \quad 
$^2$CUHK \quad
$^3$HKUST \\
$^4$Huawei Noah's Ark Lab \quad
$^5$Research Institute of Tsinghua, Pearl River Delta \\
{\tt\small wyb22@mails.tsinghua.edu.cn \qquad}
{\small $^{*}$ Equal Contribution \qquad 
\Letter~Corresponding authors} \\}

\begin{document}
\maketitle
\begin{abstract}
Current perceptive models heavily depend on resource-intensive datasets, prompting the need for innovative solutions.
Leveraging recent advances in diffusion models, synthetic data, by constructing image inputs from various annotations, proves beneficial for downstream tasks.
While prior methods have separately addressed generative and perceptive models, \methodname, for the first time, harmonizes both, tackling the challenges in generating effective data for perceptive models.
To enhance image generation with perceptive models, we introduce perception-aware loss (P.A. loss) through segmentation, improving both quality and controllability.
To boost the performance of specific perceptive models, our method customizes data augmentation by extracting and utilizing perception-aware attribute (P.A. Attr) during generation.
Experimental results from the object detection task highlight \methodname's superior performance, establishing a new state-of-the-art in layout-guided generation.
Furthermore, image syntheses from \methodname can effectively augment training data, significantly enhancing downstream detection performance. 
\end{abstract}    
\vspace{-5mm}
\section{Introduction}\label{sec:intro}

The effectiveness of current perceptive models is heavily contingent on extensive and accurately labeled datasets.
However, the acquisition of such datasets is often resource-intensive.
Recent advancements in generative models, especially diffusion models~\cite{rombach2022high}, make it possible to generate high-quality images, and thus pave the way for constructing synthetic datasets.
By providing annotations such as the class labels~\cite{rombach2022high}, segmentation maps~\cite{wang2022semantic}, and object bounding boxes~\cite{chen2023integrating}, synthetic data for generative models is proved to be useful to improve the performance on downstream tasks (\eg, classification~\cite{he2022synthetic}, object detection~\cite{chen2023integrating,bowles2018gan} and segmentation~\cite{li2023open,wu2023datasetdm}).

\begin{figure}[ht]
\centering\includegraphics[width=\linewidth]{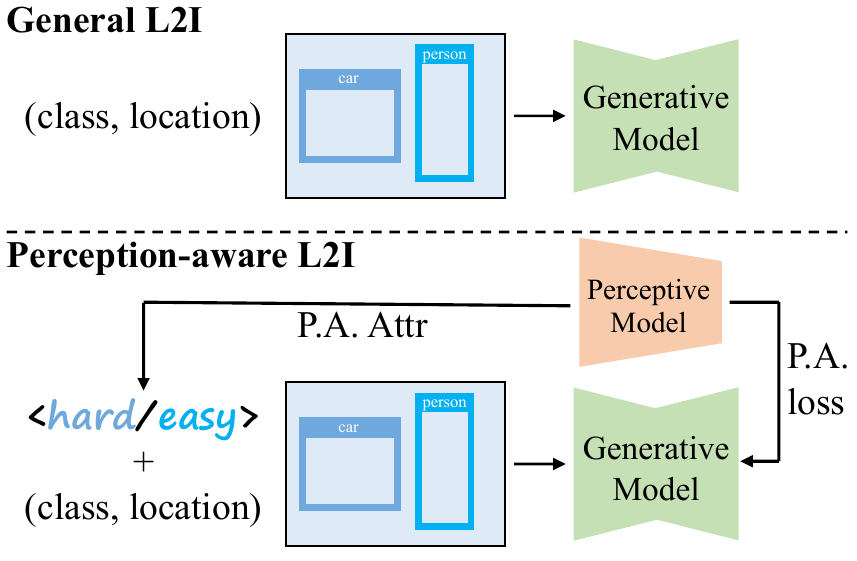}
\vspace{-7mm}
	\caption{
        \textbf{Pipeline comparison} between general L2I models (\eg, GeoDiffusion~\cite{chen2023integrating}) and our \methodname (perception-aware L2I).
        Utilizing perception-aware loss (P.A. loss) and perception-aware attributes (P.A. Attr), \methodname improves generation quality and controllability of L2I task. Perception-aware attributes further boost performance on downstream perceptive models. Moreover, perceptive model only added \textbf{1.3\% of parameters}.
        }
        \vspace{-7mm}
	\label{fig1}
\end{figure}

While most methods focus on improving generative models or perceptive models separately, the synergy between generative and perceptive models warrants a closer integration for mutual enhancement of generation and perception capabilities.
In perceptive models, the challenge lies in effective data generation or augmentation, a topic previously approached mainly from a data perspective (\eg OoD generalization~\cite{Jaipuria_2020_CVPR_Workshops,li2022coda} and domain adaptation~\cite{mullick2023domain,imbusch2022synthetic,han2021soda10m}).
\textit{Its potential to enhance perceptive models performance in general cases remains underexplored}.
Conversely, generative model research has focused on refining models for better output quality and controllability~\cite{chen2023integrating,gao2023magicdrive}. 
Nonetheless, it is essential to recognize that \textit{perceptive models can also provide valuable additional insights to assist generative models in achieving better control capabilities}.
This synergy between the generative and perceptive models offers a promising avenue for advancement, suggesting a need for more integrated approaches.

As the first work to investigate such synergy, we propose a novel perception-aware generation framework, namely \methodname, as shown in Figure~\ref{fig1}.
Our framework enables generative models to harness the information from perceptive models, thereby augmenting their capacity for controlled generation. Concurrently, it facilitates the targeted generation of data based on the capabilities of perceptive models, thereby enhancing the performance of models trained on synthetic data. Specifically, for object detection tasks, we fine-tune models based on Stable Diffusion~\cite{rombach2022high}, employing controlled generation techniques to produce high-quality data that aids in training detection models.
To elevate the quality of generation, we innovatively introduce a perception loss.
By introducing a segmentation module~\cite{cheng2022masked} based on the intermediate feature from the UNet~\cite{ronneberger2015u}, the generated content is supervised by the object mask in conjunction with label ground truth to enhance controllability.
Moreover, to further enhance the performance of detection models, we propose to extract and use object attributes from the trained detection model, and then incorporate these attributes into the training of generative models.
This approach enables the generation of new data specifically tailored to produce distinctive samples, thereby significantly improving detectors' performance.

Our experiments confirm that \methodname sets a new state-of-the-art in generation quality, achieving 31.2 mAP on COCO-Stuff. It significantly enhances detector training, increasing mAP by 0.9 mAP through the strategic use of perception-aware attribute (P.A. Attr) in training. This is largely due to \methodname's refined control in addressing long-tail data generation challenges. These advancements underscore \methodname’s technical superiority and mark a pivotal advancement in controlled image generation, especially where precise detection attributes are vital.

The main contributions of this work contain three parts:
\begin{enumerate}
    \item We propose \methodname, the first framework designed to explore the synergy between perceptive models and generative models.
    
    \item To boost generation quality, we propose a perception loss based on segmentation and object masks. To further improve the efficacy of synthetic data in perceptive models, we introduce object attributes during generation.
    
    \item Extensive experiments on object detection task show that \methodname not only achieves new SOTA in the layout-guided generation on COCO but also effectively prompts the performance for downstream detectors. 
\end{enumerate}

\begin{figure*}[ht]
	\centering
	\includegraphics[width=\linewidth]{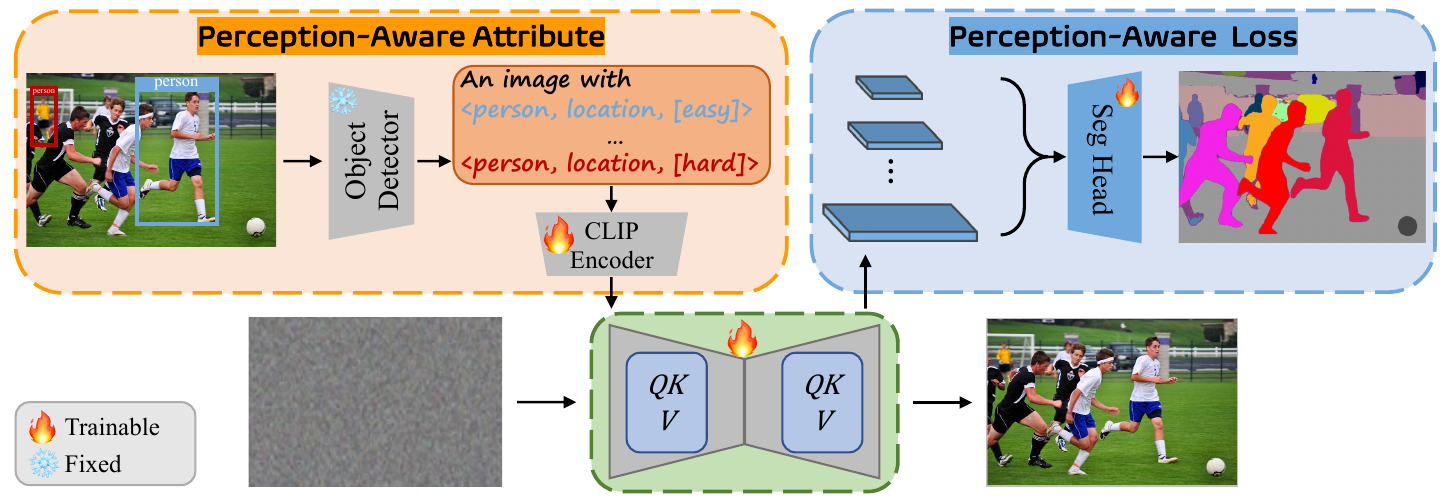}
    \vspace{-7mm}	
 \caption{\textbf{Model architecture of \methodname.} 
 To facilitate the synergy between generative models and perceptive models, we integrate two components into L2I training pipeline. Perception-aware loss (P.A. loss) leverages the segmentation head for better generation quality and controllability. Perception-aware attribute (P.A. Attr) enables \methodname to generate highly useable data for training augmentation. }
\vspace{-4mm}	
 \label{overview}
\end{figure*}
\section{Related Work}\label{sec:formatting}

\paragraph{Diffusion Models.}
Diffusion models, being one kind of generative model, are trained to learn the reverse denoising process after a forward transformation from the image distribution to the Gaussian noise distribution~\cite{ho2020denoising}.
These models can employ either a Markov process~\cite{ho2020denoising} or a non-Markov process~\cite{song2020denoising}.
Due to their adaptability and competence in managing various forms of controls~\cite{rombach2022high,zhang2023adding,li2023gligen} and multiple conditions~\cite{liu2022compositional,huang2023composer,gao2023diffguard,ma2023follow}, diffusion models have been applied in various conditional generation tasks, such as image variation~\cite{xu2023versatile}, text-to-image generation~\cite{rombach2022high}, pixel-wise controlled generation~\cite{zhang2023adding}.
A notable variation of these models is the Latent Diffusion Model (LDM~\cite{rombach2022high}).
Unlike traditional diffusion models, the LDM conducts the diffusion process in a latent space, enhancing the model's efficiency. 
Our framework for perception data generation is based on the LDM.
However, we focus on the synergy between generative models and perceptive models, proposing several designs to benefit both generation quality \& controllability and performance on downstream tasks.

\vspace{-2mm}
\paragraph{Layout-to-Image (L2I) Generation.}
Our approach focuses on converting a high-level graphical layout into a realistic image. In this context, LAMA~\cite{li2021image} implements a locality-aware mask adaptation module for improved object mask handling during image generation. Taming~\cite{jahn2021high} shows that a relatively straightforward model can surpass more complex predecessors by training in latent space. More recent developments include GLIGEN~\cite{li2023gligen}, which integrates additional gated self-attention layers into existing diffusion models for enhanced layout control, and LayoutDiffuse~\cite{cheng2023layoutdiffuse}, which employs innovative layout attention modules tailored for bounding boxes. Our generative model shares similar architecture with GeoDiffusion~\cite{chen2023integrating} and Geom-Erasing~\cite{liu2023geomerasing}, while \methodname focuses on the synergy between generation and perception, and distinctively offers
1) a novel perception-aware loss (P.A. loss) that utilizes information from the segmentation head; 2) a novel object attribute mechanism (P.A. Attr) to help the training of object detectors.

\vspace{-2mm}
\paragraph{Data Generation for Perceptive Models.}
In some L2I methods, the utility of synthetic data in enhancing object detection task performance is demonstrated, \eg, GeoDiffusion~\cite{chen2023integrating}.
Similarly, MagicDrive~\cite{gao2023magicdrive} suggests that generated images aid in 3D perception, and TrackDiffusion~\cite{li2023trackdiffusion} generates data for multi-object tracking.
However, they do not explore enhancing generation using perceptive models or tailoring data for specific detectors.
Beyond controllable generation, 
several works convert generators into perceptive models by extracting annotations from generative features.
DatasetDM \cite{wu2023datasetdm} uses a Mask2Former-style P-decoder with Stable Diffusion, while \citet{li2021image} develop a fusion module for open-vocabulary segmentation.
These techniques, while capable of producing annotated data, are limited by their reliance on text-based generation with limited annotation control, dependency on pre-trained diffusion models restricting the cross-domain applicability, and inferior performance compared to combining diffusion models with specialized models like SAM \cite{kirillov2023segment}.
\section{Method}
Our objective is to enhance the generation quality from a perceptive perspective and facilitate downstream perceptive tasks. Designing proper and strong supervision is of great importance in tackling this challenging problem, we propose to integrate easily accessible but previously neglected perceptive information i.e., perception-aware attribute (P.A. Attr) and loss (P.A. loss), into the generation framework to promote the information interaction between perceptive models and generative models. We first introduce the preliminaries in Section~\ref{latent_diffusion_model} and expand the perception-aware attribute (P.A. Attr) in detail (Section~\ref{perception-aware-attribute}), which is generated via an object detector and designed as special tokens to assist diffusion models. In Section~\ref{perception-aware-loss}, a tailored perception-aware loss (P.A. loss) is introduced. The overall architecture is depicted in Figure~\ref{overview}.

\subsection{Preliminaries}\label{latent_diffusion_model} 

Diffusion Models (DMs) have emerged as prominent text-to-image generation models, characterized by their effectiveness in generating realistic images. A notable variation, the Latent Diffusion Model (LDM)~\cite{rombach2022high}, innovatively transfers the diffusion process of standard DMs into a latent space. This transition is significant, as LDMs demonstrate the ability to maintain the original model's quality and flexibility, but with a substantially reduced computational resource requirement. This efficiency gain is primarily attributed to the reduced dimensionality of the latent space, which facilitates faster training times without compromising the generative capabilities of the model.

Stable Diffusion, an exemplary implementation of the Latent Diffusion Model (LDM), utilizes a distinctive pipeline for text-to-image (T2I) generation. The process commences with the encoding of the original image $x$ into latent space using a pre-trained Vector Quantized Variational AutoEncoder (VQ-VAE)~\cite{van2017neural}, resulting in a latent representation $z=\mathcal{E}(x)\in\mathcal{R}^{H^\prime\times W^\prime\times D^\prime}$, where is much smaller than original dimension. Concurrently, the text condition $y$ undergoes encoding via a pre-trained CLIP~\cite{radford2021learning} text encoder $\tau_\theta(\cdot)$. At a given timestep $t$, random noise is integrated into the latent variable $z$ to form $z_{t}$. The noise prediction is executed by a UNet $\epsilon_\theta(\cdot)$, which incorporates both resnet and transformer networks of varying dimensions for enhanced generative capability. The integration of the condition variables $\tau_\theta(y)$ with the UNet is achieved through cross-attention mechanisms. The formulation of the objective function can be expressed as follows:
\begin{equation}
    \mathcal{L}_{LDM} = \mathbb{E}_{\mathcal{E}(x),\epsilon\sim\mathcal{N}(0,1),t}\|\epsilon - \epsilon_\theta(z_t, t, \tau_\theta(y))\|^2.
    \label{equ:ldm}
\end{equation}
This equation represents the mean-squared error between the original noise $\epsilon$ and the noise predicted by the model, encapsulating the core learning mechanism of the Stable Diffusion model.

\subsection{Perception-Aware Attribute as Condition Input}\label{perception-aware-attribute}

To enhance the performance of detection models, this study introduces a novel approach centered around the generation of perception-aware realistic images. The methodology involves a two-step process: initially, object attributes are extracted from a pre-trained detector. These attributes encapsulate critical visual characteristics essential for accurate object detection. Subsequently, the extracted attributes are integrated into the training regime of a generative model. This integration aims to ensure that the generated images not only exhibit high realism but also align closely with the perceptive criteria crucial for effective detection. By doing so, the generative model is tailored to produce images that are more conducive to training robust detectors, potentially leading to significant improvement in detection accuracy and reliability.

\noindent\textbf{Perception-Aware Attribute.}
We define the perception-aware attribute (P.A. Attr) of an object as $d$. For each image $x$, a pre-trained detector $\mathcal{D}(\cdot)$, such as Faster R-CNN~\cite{ren2015faster} or YOLO series~\cite{bochkovskiy2020yolov4} detectors, is employed to yield $n$ predicted bounding boxes, represented as $b = [b_1,...,b_n]$ = $\mathcal{D}(x)$.  To refine the selection of bounding boxes, a filtering criterion based on a confidence score threshold $\gamma$ is applied. This process effectively retains a subset of bounding boxes that meet the threshold, resulting in a reduced set $b' = [b_1,...,b_{n'}]$, where $n'$ is significantly smaller than $n$. This selective approach ensures that only bounding boxes with a high likelihood of accurate object detection are considered, thereby enhancing the following reliability of the perception-aware attribute (P.A. Attr) extracted.

Furthermore, for each image $x$, there are $m$ ground truth objects bounding boxes, represented as $o = [o_1,...,o_m]$. The detection difficulty of each ground truth box $o_i$ is assessed based on its intersection with the $n'$ predicted boxes. Specifically, for each ground truth box, if any predicted bounding box $b_j$, where $i\in [1,...,n']$, has an intersection over union (IoU) with the ground truth box exceeding a threshold $\beta$, it is classified as $[easy]$ to detect. Conversely, the ground truth boxes without such overlapping predicted boxes are labeled as $[hard]$. This classification mechanism is encapsulated in the following expression: 
\begin{equation}
d_i =
\begin{cases}
[easy], & \text{if exists } i,  ~\text{IoU}(b_j, o_i) > \beta, \\
[hard], & \text{else}.
\end{cases}
    \label{equ:attribute}
\end{equation}

\noindent\textbf{Perception-Aware Attribute as Prompt Token.}
In this approach, each ground truth box within an image is characterized by three attributes. These include the pre-existing attributes of category ($c_i$) and location ($l_i$), along with the newly introduced attributes perception difficulty attribute ($d_i$). 
The category attribute $c_i$ is the category text itself.
For the location attribute $l_i$, the original representation is continuous coordinates. Here we discretize it via partitioning the pixel image space into a grid of location bins, and each location bin corresponds to a unique token (check more details in~\cite{chen2023integrating}).
So that we can feed the specific location token into a text encoder of L2I diffusion model~\cite{yang2023reco,chen2023integrating} to obtain the final location attribute $l_i$. 
In this way, attributes representing location, category, and ease of detection are organized into a unified representation.

Furthermore, in contrast to existing methods~\cite{yang2023reco,li2023gligen} utilizing captions as the text prompt. We design an effective text prompt equipped with multiple pairs of perception-aware attributes. Specifically, the prompt is ``\texttt{An image with \{objects\}}'', where \texttt{objects} are $[(c_1,l_1,d_1),...,(c_m,l_m,d_m)]$ and $m$ is number of ground truth bounding boxes.
This comprehensive attribute set and effective prompt aim to encapsulate a more holistic understanding of each object's characteristics, potentially providing a much richer description for perception.

\begin{figure}[t]
\centering\includegraphics[width=\linewidth]{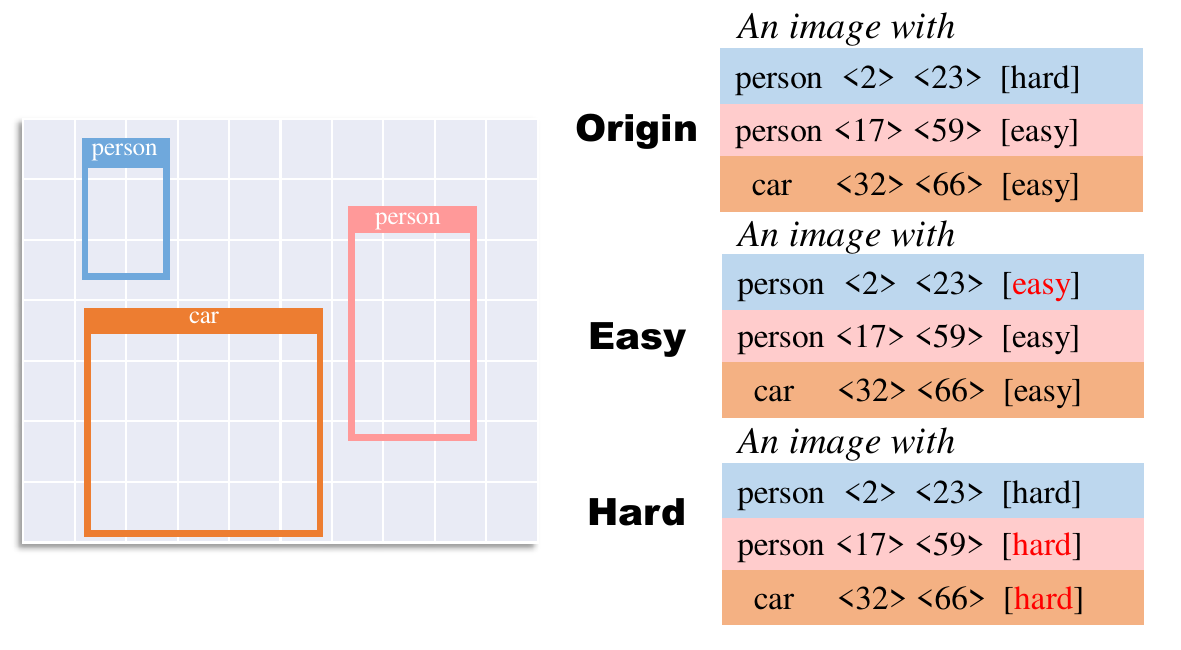}
\vspace{-5mm}
	\caption{
        \textbf{Three strategies for attribute application.} 
        Check the detailed definition in Sec.~\ref{sec:settings}.
        }
        \vspace{-4mm}
	\label{fig:three}
\end{figure}

\subsection{Perception-Aware Loss as Supervision}\label{perception-aware-loss}
During training diffusion generation model, the objective is to minimize the reconstruction distance between the predicted image (or noise) and its ground truth counterpart.
Traditional generation methodologies predominantly utilize L1 or L2 losses for this purpose.
However, these standard loss functions often fall short of producing images with high-resolution details and precise control over image attributes. To address this limitation, a novel \textit{perception-aware loss} (P.A. loss) is proposed. This loss function is constructed to leverage rich visual features, thereby facilitating more nuanced image reconstruction.

\vspace{1mm}
\noindent\textbf{Visual Features.}

Recent studies have indicated that the feature maps in the UNet model exhibit effectiveness in class-discriminative and localization tasks~\cite{wu2023diffumask,zhao2023unleashing,hertz2022prompt,wu2023datasetdm}. Inspired by these papers, we try to extract multi-scale feature maps $f = [f_1,f_2,f_3,f_4]$ from four layers of the U-Net $\epsilon_\theta(\cdot)$ of the encoding and decoding path during the training process, corresponding to resolutions of 
$(\frac{H}{8} \times \frac{W}{8}, \frac{H}{4} \times \frac{W}{4}, \frac{H}{2} \times \frac{W}{2}, H \times W )$. And we upsample $\operatorname{Upsample}(\cdot)$ and mix convolution layers $\operatorname{Conv}(\cdot)$ to concatenate the final multi-scale feature maps, which can be expressed as:
\begin{equation}
F_n=
\begin{cases} 
\operatorname{Conv}\left(f_n, \operatorname{Upsample}\left(F_{n-1}\right)\right), & n=2,3,4 \\
f_n, & n=1 
\end{cases} 
    \label{equ:feature}
\end{equation}
Thus, we have obtained the visual features $\mathcal{F} = F_4$.

\vspace{1mm}
\noindent\textbf{Perception-Aware Loss.}
To achieve a higher level of precision in image generation, this study introduces a custom-designed loss function, utilizing the rich information from perception learning.
Central to this formulation is the use of a segmentation head, which processes these features to produce instance masks, represented as $M = [m_1,..,m_k]$. In optimizing the model's high-dimensional feature space, the loss function incorporates two key components: the mask loss  $\mathcal{L}_m$ and a dice loss $\mathcal{L}_d$ following~\cite{wu2023datasetdm}. 
The integration of these losses, specifically tailored for perceptive enhancement, serves to finely tune the U-Net’s capabilities. This approach allows for more granular control over the generated images, leveraging the perceptive information embedded within the features for more accurate image synthesis.

Considering that noise has been added to the visual features $\mathcal{F}$, we assign different scales to the loss term, easing the burden of model training.
Specifically, we adopt $\bar{\alpha}_t$ from Denoising Diffusion Probabilistic Models (DDPM), as delineated in \citet{ho2020denoising}, to mitigate this effect. This scheme is designed to counterbalance the noise component in the features, ensuring the integrity and utility of $\mathcal{F}$.
Given the DDPM scheduler adds Gaussian noise to the data according to a variance schedule $\beta_1, \ldots, \beta_t$, and then using the notation $\alpha_{t} := 1 - \beta_t$ and $\bar{\alpha}_t := \prod_{s=1}^t \alpha_s$.
The perception-aware loss (P.A. loss) can be expressed as:
\begin{equation}
\mathcal{L}_p = \sqrt{\bar{\alpha}_t}(\mathcal{L}_m + \mathcal{L}_d),
    \label{equ:per_loss}
\end{equation}
where the $\sqrt{\bar{\alpha}_t}$ is designed to reduce the impact of feature maps with higher noise levels, thereby emphasizing feature maps with lower noise (\ie, smaller time steps).

\vspace{1mm}
\noindent\textbf{Objective Function.}
Ultimately, our objective function combines the perception-aware loss with the foundational loss function of the Latent Diffusion Model (LDM). This integration is mathematically represented as follows:
\begin{equation}
\mathcal{L} = \mathcal{L}_{LDM} + \lambda\mathcal{L}_p
    \label{equ:total loss}
\end{equation}

For the purposes of this model $\lambda$ is set to 0.01, ensuring a balanced incorporation of the perception-aware components while maintaining the primary structure and goals of the LDM loss function. This calibrated approach allows for a nuanced optimization that leverages the strengths of both losses, thereby enhancing the model's performance in generating high-quality, perception-aligned images.
\section{Experiments}
\subsection{Experiment Settings}\label{sec:settings}
\paragraph{Dataset.}
We employ the widely recognized COCO-Thing-Stuff benchmark~\cite{caesar2018coco,lin2014microsoft,li2021image} for the L2I task, which includes 118,287 training images and 5,000 validation images. Each image is annotated with bounding boxes and pixel-level segmentation masks for 80 categories of objects and 91 categories of stuff. In line with previous works~\cite{cheng2023layoutdiffuse,yang2023reco,chen2023integrating}, we ignore objects belonging to crowds or occupying less than 2\% of the image area.

\begin{table}[t!]
\centering
\setlength{\tabcolsep}{1mm}
\begin{tabular}{lc|c|ccc}
\toprule
Method & Epoch & FID$\downarrow$ & mAP$\uparrow$ & AP$_{50}$$\uparrow$ & AP$_{75}$$\uparrow$ \\
\midrule
LostGAN & 200 & 42.55 & 9.1 & 15.3 & 9.8 \\
LAMA & 200 & 31.12 & 13.4 & 19.7 & 14.9 \\
TwFA & 300 & 22.15 & - & 28.2 & 20.1 \\
\midrule
Frido & 200 & 37.14 & 17.2 & - & -  \\
L.Diffuse$^\dag$ & 60 & 22.20 & 11.4 & 23.1 & 10.1 \\
L.Diffusion$^\dag$ & 180 & 22.65 & 14.9 & 27.5 & 14.9 \\
ReCo$^\dag$ & 100 & 29.69 & 18.8 & 33.5  & 19.7 \\
GLIGEN & 86 & 21.04 & 22.4 & 36.5 & 24.1 \\
ControlNet$^\dag$ & 60 & 20.37 & 24.8 & 36.6 & 27.7 \\
GeoDiffusion & 60 & 20.16 & 29.1 & \uit{38.9} & 33.6 \\
\midrule
\rowcolor{backcolor}
\methodname$_{origin}$ & 60 & \textbf{19.28} & \uit{29.8} & 38.6 & \uit{34.1} \\
\rowcolor{backcolor}
\methodname$_{hard}$ & 60 & 19.72 & 25.3 & 33.7 & 29.1 \\
\rowcolor{backcolor}
\textbf{\methodname$_{easy}$} & 60 & \uit{19.66} & \textbf{31.2} & \textbf{40.2} & \textbf{35.6} \\
\bottomrule
\end{tabular}
\vspace{-2mm}
\caption{Evaluation of image quality and correspondence to layout on COCO val-set. The best results are in \textbf{bold} and the second best results are \uit{underlined italic}. $^\dagger$~Implemented by ourselves.}
\label{tab:fidelity}
\vspace{-5mm}
\end{table}

\noindent\textbf{Implementation Details.}
We fine-tune \methodname from the Stable Diffusion v1.5~\cite{patil2022stable} checkpoint. We introduce location tokens into the text encoder and initialize the embedding matrix of the location tokens with 2D sine-cosine embedding. With the VQ-VAE~\cite{van2017neural} fixed, we fine-tune all parameters of the text encoder and use AdamW~\cite{loshchilov2017decoupled} optimizer with a cosine learning rate schedule of $1e^{-4}$. And the linear warm-up is adopted in the first 3000 steps. The text prompt is replaced with a null text for unconditional generation with a probability of 10\%. The model is trained on 8$\times$32GB GPUs with a batch size of 32, requiring about 20 hours for 60 epochs. We sample images using DPM-Solver~\cite{lu2022dpm} scheduler for 50 steps with CFG at 3.5.

\noindent\textbf{Strategy for Attribute Application.}
Upon completion of the training process, it is flexible to apply perception-aware attribute (P.A. Attr) during the generation.
For the purpose of simple yet effective validation, we adopt three attribute strategies in Fig~\ref{fig:three}: 1) \methodname$_{origin}$: the original perception-aware attribute (P.A. Attr). We obtain the attributes of each object in the image using a detector and use them directly for generation. 2) \methodname$_{hard}$: all objects are assigned the $[hard]$ attribute. All objects are treated as difficult samples for perception. 3) \methodname$_{easy}$: all objects are assigned the ${easy}$ attribute. All objects are treated as easy samples for perception.

\subsection{Main Results}
The L2I generation requires the generated objects to be as consistent as possible with the original image while ensuring high-quality image generation. Therefore, we will first comprehensively analyze the fidelity experiment in Section~\ref{sec:fidelity}. Additionally, an important purpose of generating target detection data is its applicability to downstream target detection. We present the trainability experiment in Section~\ref{sec:trainability}.

\begin{figure*}[t]
	\centering
	\includegraphics[width=\linewidth]{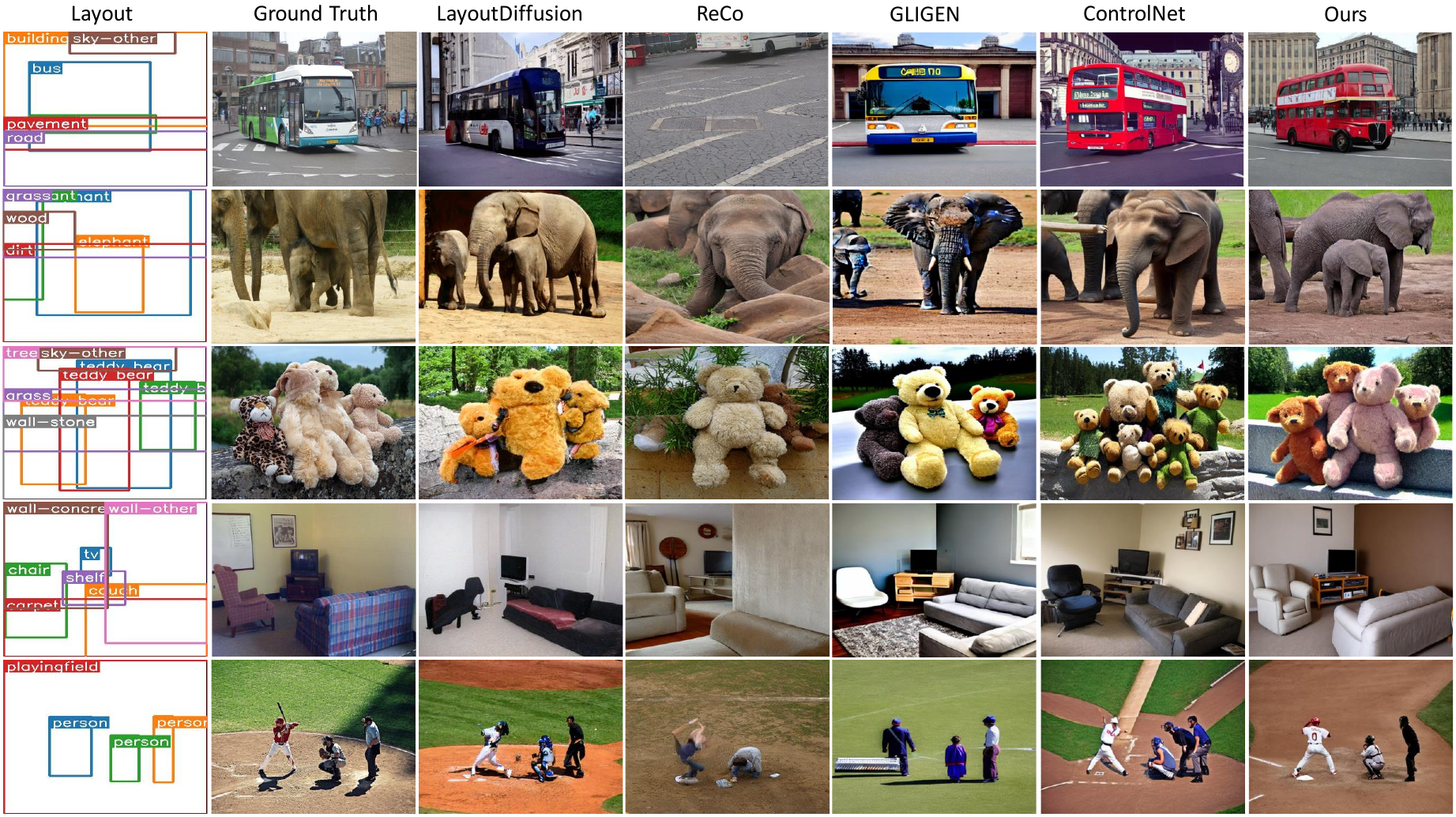}
    \vspace{-7mm}
    \caption{
    \textbf{Qualitative comparison on the Microsoft COCO dataset.}
    Our \textit{DetDiffusion} can generate highly realistic images consistent with the provided semantic layouts.
    }
    \vspace{-4mm}
    \label{qualit}
\end{figure*}

\subsubsection{Fidelity}\label{sec:fidelity}
\noindent\textbf{Set up.}
To evaluate fidelity, we utilize two primary metrics on the COCO-Thing-Stuff validation set. The Fréchet Inception Distance (FID)~\cite{heusel2017gans} assesses the overall visual quality of the generated image. It measures the distinction in feature distribution between the real images and the generated images using an ImageNet-pretrained Inception-V3~\cite{szegedy2016rethinking} network.
The YOLO Score~\cite{li2021image} in LAMA~\cite{li2021image} uses the mean average precision (mAP) of 80 object categories' bounding boxes on generated images. It is achieved using a pre-trained YOLOv4~\cite{bochkovskiy2020yolov4} model, demonstrating the precision of object detection in a generated model. Our model is trained on the image size of 256$\times$256. Following previous work, we utilize images containing between 3 to 8 objects, resulting in 3,097 images during validation.

\noindent\textbf{Results.}
we evaluated our models with three attribute strategies on the COCO-Thing-Stuff validation set and compared them with state-of-the-art models for L2I task such as LostGAN~\cite{sun2019image}, LAMA~\cite{li2021image}, TwFA~\cite{yang2022modeling}, Frido~\cite{fan2023frido}, LayoutDIffuse~\cite{cheng2023layoutdiffuse}, LayoutDiffusion~\cite{zheng2023layoutdiffusion}, Reco~\cite{yang2023reco}, GLIGEN~\cite{li2023gligen}, GeoDiffusion~\cite{chen2023integrating}, and ControlNet~\cite{zhang2023adding}. As shown in Table~\ref{tab:fidelity}, our \methodname$_{origin}$ and \methodname$_{easy}$ strategies outperformed all competitors across all metrics. The \methodname$_{origin}$ strategy achieved the best FID (19.28) and outperformed other models in YOLO Score. This indicates that using perception-aware loss (P.A. loss) and treating the perception-aware attribute (P.A. Attr) as an additional condition obtained from perception can generate more realistic images.
The \methodname$_{easy}$ strategy achieved a YOLO Score exceeding the best model by 2.1mAP, and showed significant improvement compared to the \methodname$_{origin}$ strategy that demonstrates generated examples are easy for the detector to perceive.
The \methodname$_{hard}$ strategy is designed to generate examples that are more challenging for the detector, and the results are in line with our expectations. The YOLO Score decreased compared to the \methodname$_{origin}$ strategy, indicating that generated examples are more difficult. The significance of the \methodname$_{hard}$ strategy lies in its impact on trainability, which is demonstrated in Section~\ref{sec:trainability}.

The enhanced FID and YOLO Score achieved with the \methodname$_{origin}$ approach illustrates the effectiveness of incorporating P.A. Attr and P.A. loss in regulating the performance of the images. Furthermore, the \methodname$_{easy}$ and \methodname$_{hard}$ strategies demonstrate our model's ability to comprehend and manipulate the attributes of $[easy]$ and $[hard]$ from the perceptual models, thus enabling control over the difficulty level.

\begin{figure}[t]
	\centering
	\includegraphics[width=\linewidth]{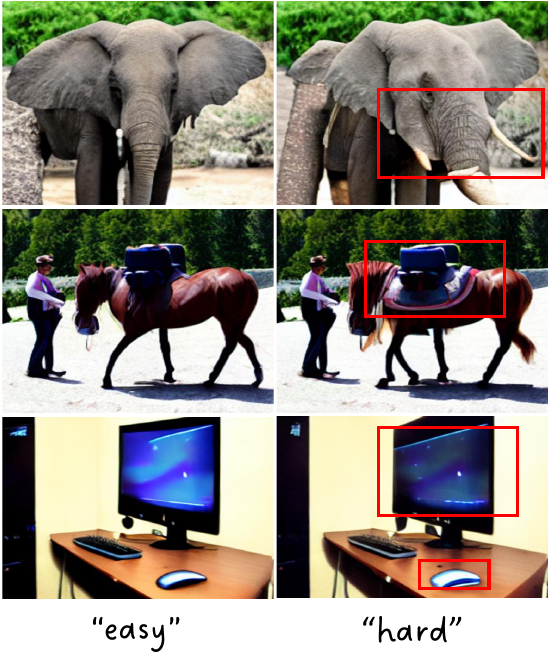}
\vspace{-6mm}	
 \caption{
 \textbf{Qualitative comparison on the perception-aware attribute.}
 Although provided with the exact same semantic layouts, simply changing the perception-aware attribute (P.A. Attr) among $[easy]$ (left) and $ [hard]$ (right) can effectively alter the low-level image pattern of generated images.
 The former achieves better detector recognizability, while the latter performs as better augmentation samples, as demonstrated in Table~\ref{tab:fidelity} and~\ref{tab:trainability} respectively.
 }
 \vspace{-7mm}
 \label{ab:easy}
\end{figure}

\subsubsection{Trainability}\label{sec:trainability}
\noindent\textbf{Set up.}
This section explores the potential advantages of using generated images from \methodname for training object detectors. The evaluation of trainability includes using a pre-trained L2I model to create a new synthetic training set from the original annotations. Both the original and synthetic training sets are then employed to train a detector.

\vspace{1mm}
\noindent\textbf{COCO Trainability.}
To establish a reliable baseline, we utilize the COCO2017 dataset, selectively choosing images containing 3 to 8 objects to enhance synthetic image quality and maintain fidelity. This process yielded a training set comprising 47,429 images with 210,893 objects. The goal is to showcase the improvements \methodname can contribute to downstream tasks, maintaining fixed annotations for various model comparisons. For training efficiency and focused evaluation of data quality's impact on training, we adopt a modified 1× schedule, reducing the training period to 6 epochs. \methodname is trained on images resized to 800$\times$456, its maximum supported resolution, to reconcile resolution differences with COCO.

\begin{table}[t!]
\centering
\setlength{\tabcolsep}{1mm}
\begin{tabular}{l|c|cc|cc}
\toprule
Method & mAP & AP$_{50}$ & AP$_{75}$ & AP$^{m}$ & AP$^{l}$ \\
\midrule
Real only & 34.5 & 55.5 & 37.1 & 37.9 & 44.3 \\ 
\midrule
L.Diffusion & 34.0 & 54.5 & 36.5 & 37.2 & 43.6 \\
GLIGEN & 34.3 & 54.8 & 36.7 & 37.4 & 44.3 \\
ControlNet & 34.4 & 54.5 & 36.9 & 37.8 & 45.0 \\
ReCo & 33.6 & 53.2 & 36.2 & 36.7 & 44.0 \\
GeoDiffusion & 34.8 & 55.3 & 37.4 & 38.2 & 45.4 \\
\midrule
\rowcolor{backcolor}
\methodname$_{origin}$ & \uit{35.3} & \uit{55.7} & \uit{38.2} & \uit{38.4} & \uit{46.5} \\
\rowcolor{backcolor}
\textbf{\methodname$_{hard}$} & \textbf{35.4} & \textbf{55.8} & \textbf{38.3} & \textbf{38.5} & \textbf{46.6} \\
\rowcolor{backcolor}
\methodname$_{easy}$ & 35.2 & 55.5 & 37.9 & 38.3 & 46.3 \\
\bottomrule
\end{tabular}
\vspace{-2mm}
\caption{
Comparison of trainability on COCO. \methodname leads to better improvements by emphasizing hard objects in augmentation. The best results are in \textbf{bold} and the second best results are \uit{underlined italic}.
}
\label{tab:trainability}
\end{table}

\noindent\textbf{Results.}
As shown in Table~\ref{tab:trainability}, ReCO~\cite{yang2023reco}, GeoDiffusion~\cite{chen2023integrating}, and all our three strategies can be beneficial for the training of downstream detectors, with the synthetic images generated by the strategies showing a more significant gain for the detector (exceeding 35.0 mAP). Furthermore, compared to the ``origin'' strategy, the ``hard'' strategy exhibits the most improvement across all detector metrics. This is attributed to the generation of more challenging instances through the ``hard'' strategy, which often represents the long-tail data in real datasets or serves as a form of stronger data augmentation. Overall, our model's generated data significantly enhances the training of downstream detectors, surpassing all other L2I models, and reveals that information obtained through perception can further benefit downstream training.

\begin{figure}[t!]
  \begin{subfigure}{0.5\linewidth}
    \centering
    \includegraphics[width=\linewidth]{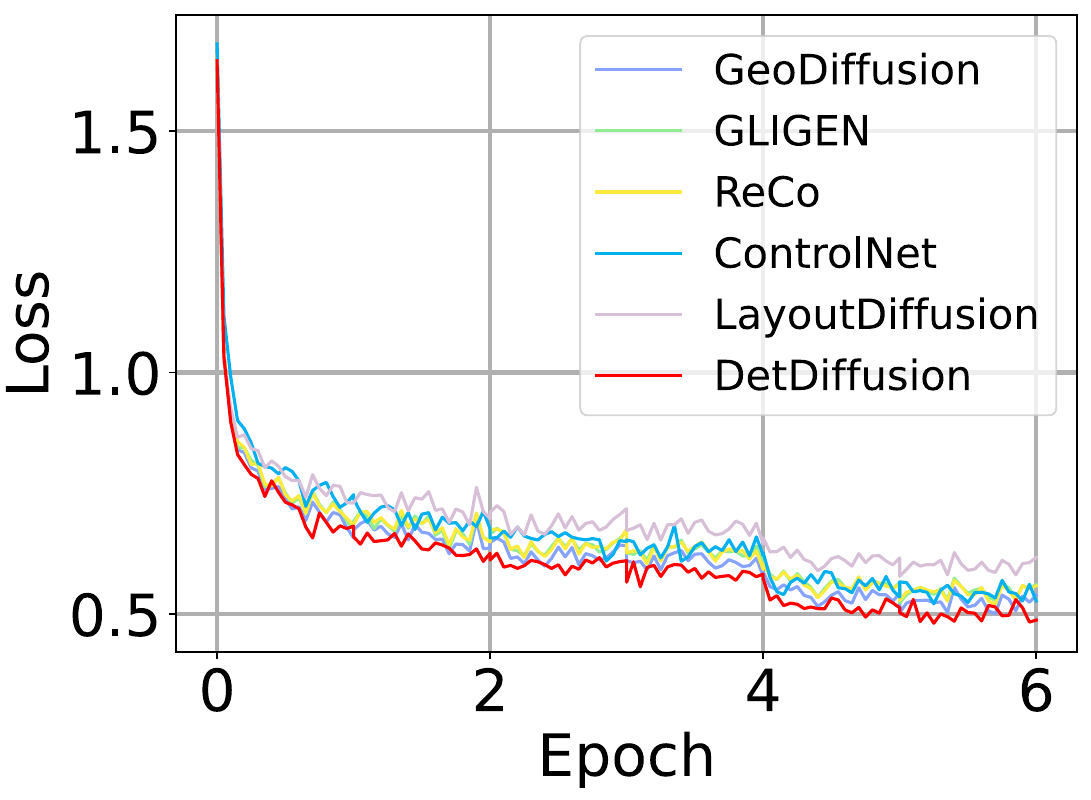}
    \caption{Training loss curve.}
    \label{fig:subfig1}
  \end{subfigure}%
  \begin{subfigure}{0.5\linewidth}
    \centering
    \includegraphics[width=\linewidth]{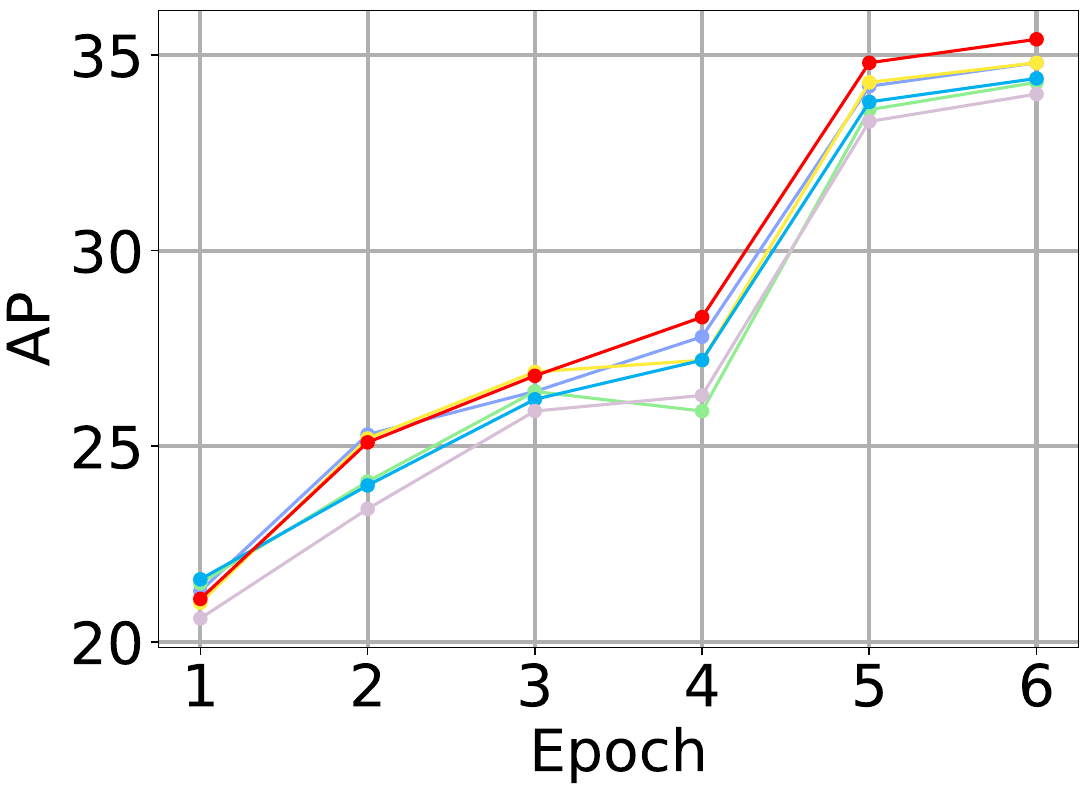}
    \caption{Validation mAP curve.}
    \label{fig:subfig2}
  \end{subfigure}
  \vspace{-7mm}
  \caption{\textbf{Training loss and validation mAP curves.}}
  \label{fig:total}
\vspace{-6mm}
\end{figure}

To verify the \textit{training effectiveness with equal training costs}, we plot the training loss curve and val mAP curve in Figure \ref{fig:subfig1} and \ref{fig:subfig2} respectively.
Our \methodname achieves the best performance throughout the training procedure.

We present more results on trainability in Table~\ref{tab:more trainability}, focusing on \textbf{less frequent categories} in the COCO dataset such as parking meter, scissor and microwave, each accounting for less than 0.2\% of the dataset. It can be observed that our hard strategy yields gains across all categories, with particularly significant improvements for long-tail categories.

\begin{table*}[t!]
\setlength{\tabcolsep}{1.35mm}
\begin{center}
\begin{tabular}{l|c|ccccccccccc}
\toprule
\multirow{2}*{Method}  & \multicolumn{9}{c}{Average Precision$\uparrow$} \\
\cline{2-13}
 & mAP &  parking. & scissor & micro. & mouse & keyboard & hot dog & baseball. & sandwich & train & skateboard & tv  \\
\midrule
real only  & 34.5 & 41.6 & 17.8 & 49.2 & 55.6 & 43.8 & 23.6 & 31.3 & 29.8 & 51.9 & 41.7 & 49.8  \\
\midrule
easy &  35.2 & 42.5 & 17.8 & 47.3 & 54.5 & 45.0 & 24.6 & 32.0 & 30.3 & 53.8 & 41.9 & 50.4 \\
orgin &  35.3 & 40.4 & 17.7 & 51.5 & 55.5 & 44.3 & 25.4 & 30.9 & 30.2 & 53.8 & 43.4 & 50.2 \\
\rowcolor{backcolor}
\textbf{hard} & \textbf{35.4} & \textbf{43.3} & \textbf{19.4} & \textbf{53.2} & \textbf{56.3} & \textbf{45.5} & \textbf{25.4} & \textbf{32.7} & \textbf{31.4} & \textbf{54.1} & \textbf{44.1} & \textbf{50.6} \\
\bottomrule
\end{tabular}
\end{center}
\vspace{-6mm}
\caption{\textbf{Rare categories results of trainability on COCO2017.} The best results are in \textbf{bold}.
 The ``parking.'', ``micro.'' and ``baseball.'' suggest parking meter, microwave and baseball gloves.}
\vspace{-5mm}
\label{tab:more trainability}
\end{table*}

\begin{table}[t!]
\captionsetup{width=.95\linewidth}
\setlength{\tabcolsep}{4mm}
\begin{center}
\begin{tabular}{cc|cc}
\toprule
P.A. Attr & P.A. loss & FID$\downarrow$ & mAP$\uparrow$ \\
\midrule
 &  & 20.16 & 29.1 \\
\checkmark & & 19.92 & 30.4 \\
\rowcolor{baselinecolor}
\checkmark & \checkmark & \textbf{19.66} & \textbf{31.2} \\
\bottomrule
\end{tabular}
\vspace{-2mm}
\caption{{\bf Ablations on essential components of \textit{DetDiffusion}
} for perception-awareness.
Best results are achieved when both components are adopted.
}
\label{tab:ablation}
\vspace{-8mm}
\end{center}
\end{table}

\begin{table}[t!]
\captionsetup{width=\linewidth}
\setlength{\tabcolsep}{2mm}
\small
\begin{center}
\begin{tabular}{cc|ccc}
\toprule
Detector & Method & mAP$\uparrow$ & AP$_{50}$$\uparrow$ & AP$_{75}$$\uparrow$ \\
\midrule
\multirow{4}{*}{FCOS} & Real only& 33.7 & 52.9 & 35.7 \\
& \methodname$_{origin}$& 34.9 & 53.8 & \textbf{37.0} \\
& \methodname$_{easy}$& 34.8 & 53.8 & 36.7 \\
& \textbf{\methodname$_{hard}$}& \textbf{35.0} & \textbf{54.0} & 36.9 \\
\midrule
\multirow{4}{*}{ATSS} & Real only& 36.3 & 53.9 & 39.1 \\
& \methodname$_{origin}$& 37.2 & 54.8 & 40.2 \\
& \methodname$_{easy}$& 37.1 & 54.6 & 40.0 \\
& \textbf{\methodname$_{hard}$}& \textbf{37.4} & \textbf{55.0} & \textbf{40.5} \\
\bottomrule
\end{tabular}
\vspace{-2mm}
\caption{{\bf Trainability for more detectors on COCO.} 
}
\vspace{-8mm}
\label{tab:more detector}
\end{center}
\end{table}

\subsection{Qualitative Results}
\noindent\textbf{Fidelity.}
Figure~\ref{qualit} displays examples that validate our model's fidelity and accuracy in image generation. LayoutDiffusion's chaotic results stem from its extra control modules clashing with the diffusion process. ReCo, dependent on high-quality captions, often suffers quality reduction and misses details. GLlGEN and ControlNet, despite high-quality outputs, lack precise object supervision, leading to insufficient detail and variable object quantities. Our implementation of P.A. loss and P.A. Attr enhances object quality, ensuring consistent quantities and controlled generation, as reflected in the alignment of generated object numbers with P.A. Attr.

\vspace{-2mm}
\noindent\textbf{Easy and Hard.}
In Figure~\ref{ab:easy}, we present perception-aware attribute (P.A. Attr) selections, comparing "easy" and "hard" instances. The "easy" images, exemplified by elephants, horses, monitors, and keyboards, are generated with an emphasis on intrinsic object features, ensuring clarity and lack of noise. Conversely, the "hard" examples, such as elephants with tusks, saddled horses, dim monitors, and reflective mice, incorporate additional elements that introduce noise through occlusions, lighting, and other complexities. These attributes make object recognition more challenging. Notably, there are both clearly distinguishable and subtly different "easy" and "hard" cases, highlighting the nuanced impact on the detection process. This indicates the identification of challenging examples without prior knowledge. For further illustrations, see Appendix~\ref{appde:quali}.

\begin{table}[t!]
\captionsetup{width=\linewidth}
\setlength{\tabcolsep}{2.5mm}
\small
\begin{center}
\begin{tabular}{c|c|ccc}
\toprule
 &  FID$\downarrow$ & mAP$\uparrow$ & AP$_{50}$$\uparrow$ & AP$_{75}$$\uparrow$ \\
\midrule
Faster R-CNN & 19.99 & 29.5 & 39.2 & 33.8 \\
\rowcolor{baselinecolor}
\textbf{YOLOv4}  & \textbf{19.92} & \textbf{30.4} & \textbf{40.8} & \textbf{35.1}\\
\bottomrule
\end{tabular}
\vspace{-2mm}
\caption{{\bf Perception aware attribute from different detectors.} 
Note the results are both evaluated with YOLOv4.
Better performance is achieved if the perception-aware attribute is provided by the evaluated detector specifically.
}
\label{tab:ab detector}
\vspace{-8mm}
\end{center}
\end{table}

\subsection{Ablation Study}
\noindent\textbf{Model components.}
We sequentially integrate two modules into the baseline model to evaluate our model's key elements. For a clear demonstration of P.A. loss effects, all attributes are set as $[easy]$. As Table~\ref{tab:ablation} indicates, adding P.A. Attr notably enhances image fidelity and YOLO Score. This implies that perceptual information inclusion aids in producing more realistic and recognizable images. Furthermore, implementing P.A. loss, which oversees potential features in intermediately generated images, significantly improves the model's precision in image generation, especially in positional accuracy.

\noindent\textbf{Trainability.}
We further conduct experiments on FCOS and ATSS.
As shown in Table~\ref{tab:more detector}, images generated by \methodname achieve significant improvement regardless of the detector models, consistently with results in Table~\ref{tab:trainability}.

\noindent\textbf{Detector.}
We explore two widely recognized detectors~\cite{ren2015faster,bochkovskiy2020yolov4} for acquiring P.A. Attr in experiments that omit the use of P.A. loss. Table~\ref{tab:ab detector} demonstrates the detector choice significantly affects P.A. Attr quality, with YOLOv4 outperforming in this aspect. Therefore, YOLOv4 serves as a primary detector for Fidelity, while Faster R-CNN is used for trainability due to its role as a trained downstream detector.

\section{Conclusion}
This paper proposes \methodname, a simple yet effective architecture to utilize the intrinsic synergy between generative and perceptive models.
By incorporating detector-awareness into geometric-aware diffusion models via P.A. Attr as conditional inputs and P.A. loss as supervision, 
\textit{DetDiffusion} can generate detector-customized images for better recognizability and trainability.

\vspace{-4mm}
\paragraph{Acknowledgments.}
This work is supported by the Project from Science and Technology Innovation Committee of Shenzhen (KCXST20221021111201002) and the key-Area Research and Development Program of Guangdong Province (2020B0909050003).
We also acknowledge the support of MindSpore, CANN, and Ascend AI Processor used for this research. 
This research is supported by the Research Grants Council of Hong Kong through the Research Impact Fund project R6003-21 and the Research Matching Grant Scheme under Grant No. 8601440.
\clearpage
{
    \small
    \bibliographystyle{ieeenat_fullname}
    \bibliography{main}
}
\clearpage
\maketitlesupplementary
\appendix

\section{Details of Experiments}

\subsection{Fidelity}
The experiments on Fidelity are conducted using the COCO-Thing-Stuff dataset~\cite{caesar2018coco}, and its perception-aware attribute is derived from a pre-trained YOLOv4. As the detector can only identify the 80 categories in COCO2017~\cite{lin2014microsoft}, objects that cannot be detected are assigned the perception aware attribute of [background]. For example, the obtained text prompt is "\texttt{An image with (person,<23><44>,[easy]),(person,<45>
<80>,[hard]),(playingfield,<0><400>,
[background])}", where the location token follows the approach~\cite{chen2023integrating}, using two location bins to represent the upper-left and lower-right coordinates of the object.

\vspace{-1mm}
\subsection{Trainability}
We conducted training of our \methodname model on the COCO-Thing-Stuff dataset at a resolution of 800x456. The perception-aware attribute of COCO-Thing-Stuff is derived from a pre-trained Faster R-CNN~\cite{ren2015faster}.

We then employ the trained generative model to generate a subset of the COCO2017 training set, comprising 47,429 images, using three distinct strategies: \methodname$_{easy}$,\methodname$_{hard}$, and \methodname$_{origin}$. In this context, \methodname$_{origin}$ is also obtained through detection utilizing Faster R-CNN on the 47,429 images, with the corresponding detection outcomes presented in Table~\ref{tab:ab detector}. Finally, a Faster R-CNN with an R-50-FPN[mmdetection] backbone was trained using the combined 47,429 images and the coco2017 training set, followed by an evaluation of its performance on the coco2017 validation set.

\vspace{-1mm}
\section{More Results of Experiments}
\subsection{Ablation on perception-aware loss}
This section investigates two important components of the perception aware loss, namely $\sqrt{\bar{\alpha}_t}$~\cite{ho2020denoising} and dice loss. Table~\ref{tab:ab PAloss} shows that $\sqrt{\bar{\alpha}_t}$ is crucial for perception aware loss as it can reduce the impact of noise to some extent. And dice loss is also essential as a complement to mask loss.

\begin{table}[t!]
\captionsetup{width=\linewidth}
\setlength{\tabcolsep}{2.5mm}
\small
\begin{center}
\begin{tabular}{c|ccc}
\toprule
  & mAP$\uparrow$ & AP$_{50}$$\uparrow$ & AP$_{75}$$\uparrow$ \\
\midrule
\rowcolor{baselinecolor}
\textbf{\methodname} & \textbf{31.2} & \textbf{40.2} & \textbf{35.6}  \\
w/o $\sqrt{\bar{\alpha}_t}$ & 29.6 & 38.6 & 34.3 \\
w/o dice loss & 29.6 & 39.4 & 34.5 \\
\bottomrule
\end{tabular}
\caption{{\bf Ablation on perception-aware loss.} We display the ablation experiments about $\sqrt{\bar{\alpha}_t}$ and dice loss.
}
\vspace{-1cm}
\label{tab:ab PAloss}
\end{center}
\end{table}

\vspace{-0.5mm}
\section{More Discussion}
\paragraph{Limitation.} 
Currently, images generated by \methodname can only be utilized to train object detectors.
More flexible usage of generated images including the incorporation with the generative pre-training~\cite{chen2023mixed,zhili2023task} and the contrastive learning~\cite{chen2021multisiam,liu2022task} is an interesting future research direction.
How to generate high-quality images aligned with human values without harmful and toxic content~\cite{chen2023gaining,schramowski2023safe,gou2023mixture} is also important for the practical usage of \textit{DetDiffusion}.

\section{More Qualitative Results}
\label{appde:quali}

\begin{figure*}[t]
	\centering
	\includegraphics[width=\linewidth]{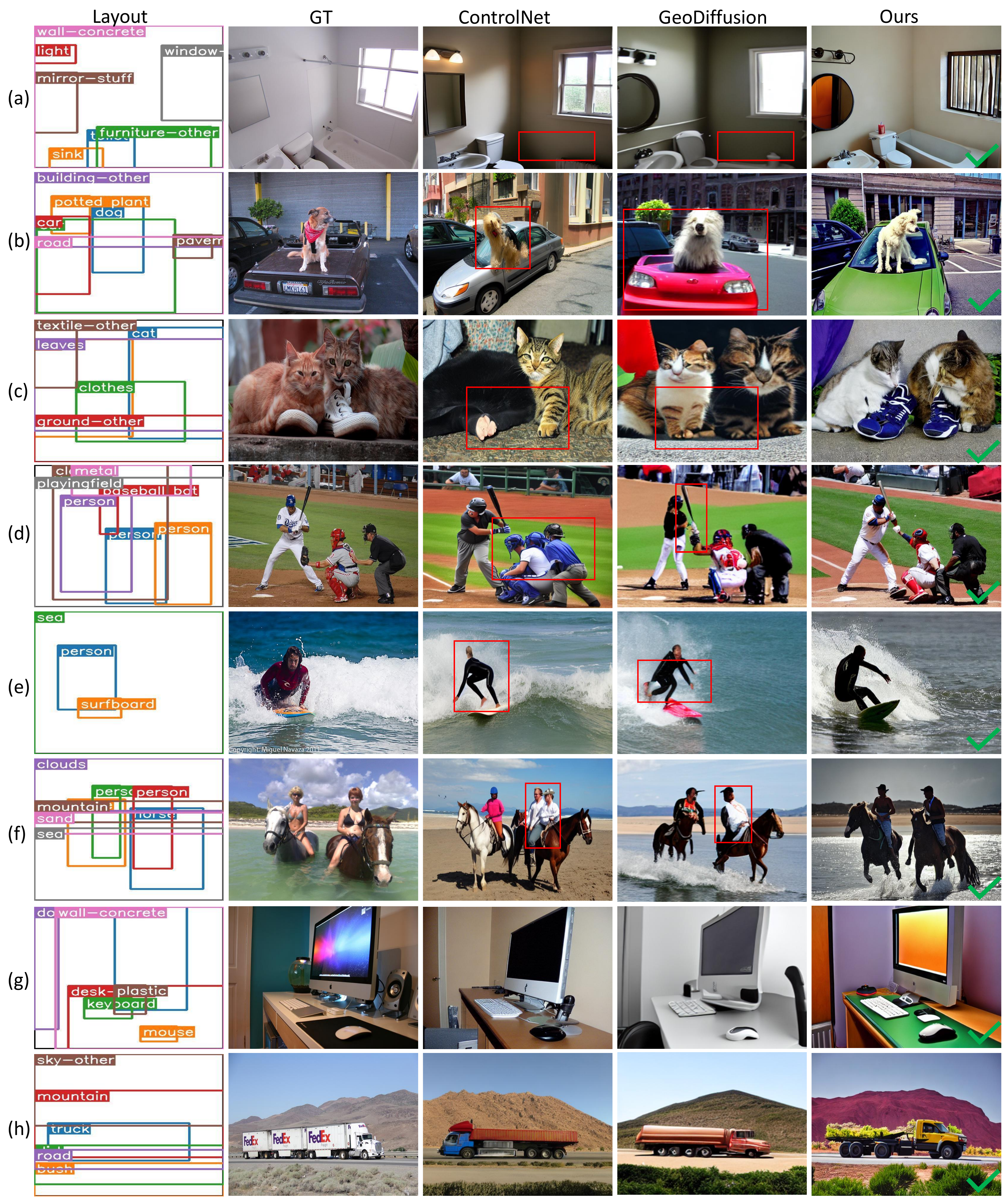}
    \caption{
    \textbf{More qualitative comparison on the COCO dataset.} We highlight some region with \textcolor{red}{red boxes} to facilitate comparison.
    }
    \label{qualit more}
\end{figure*}

Figure~\ref{qualit more} presents a comparison between our generated images and some state-of-the-art models~\cite{zhang2023adding,chen2023integrating}. Our results demonstrate accurate hierarchical relationships, as evidenced by the realistic depiction of objects such as the car and dog, and the high realism of dynamic human figures. Additionally, our generated images exhibit high quality, as illustrated in (g) and (h).

\begin{figure*}[t]
	\centering
	\includegraphics[width=\linewidth]{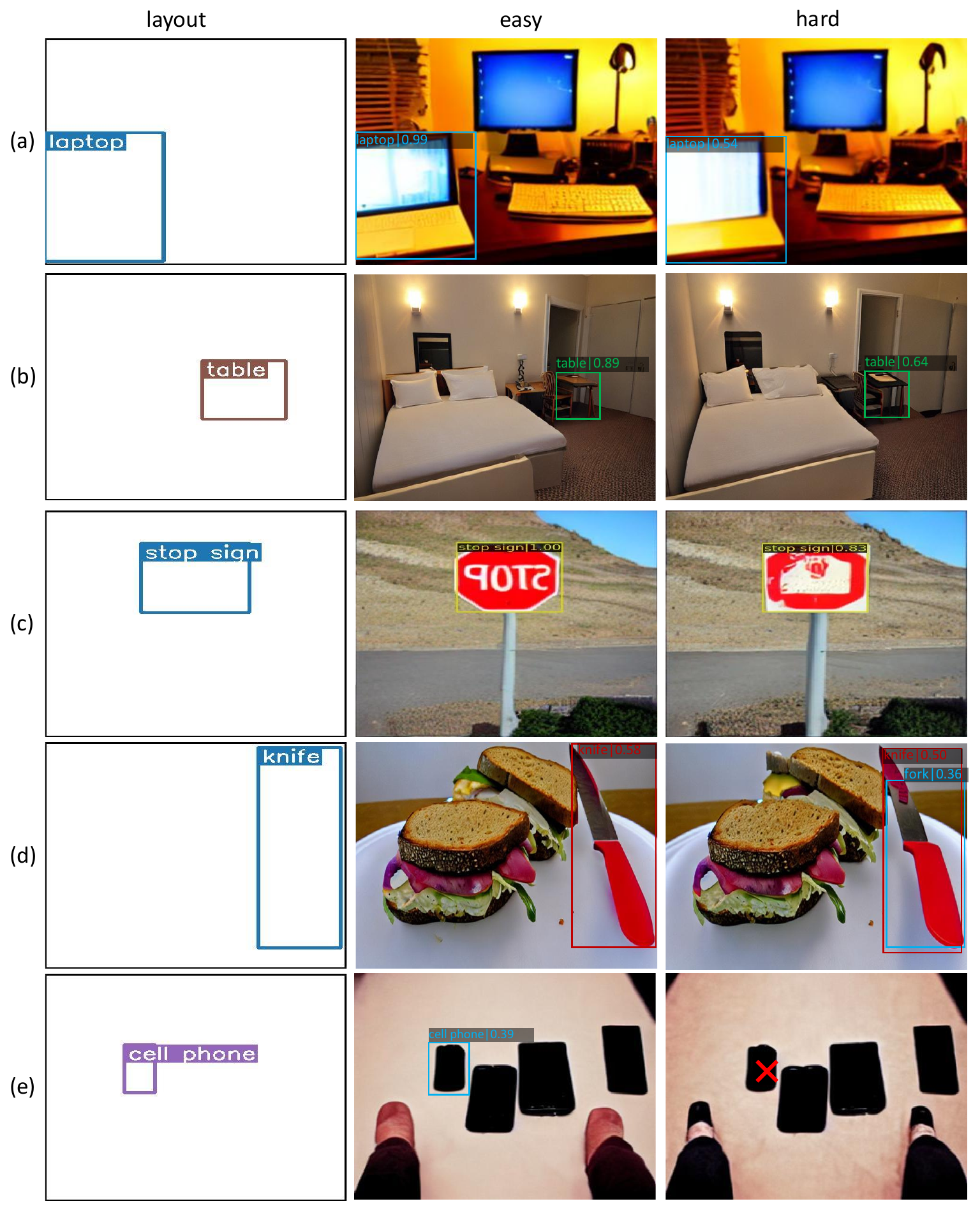}
    \caption{
    \textbf{More examples of easy and hard perception-aware attribute.} These examples are apparent to see the gap.
    }
    \vspace{-4mm}
    \label{easy hard sup 1}
\end{figure*}

\begin{figure*}[t]
	\centering
	\includegraphics[width=\linewidth]{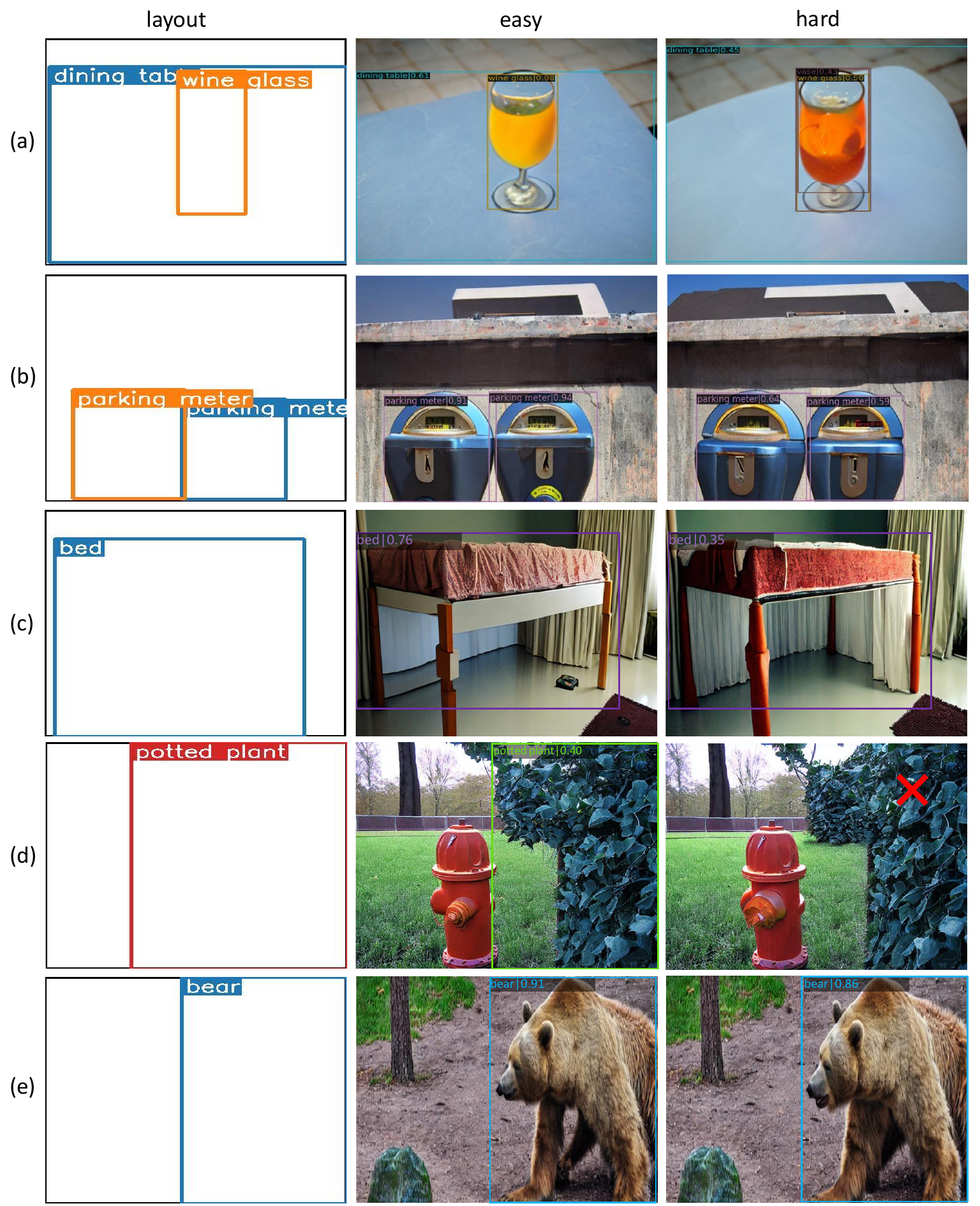}
    \caption{
    \textbf{More examples of easy and hard perception-aware attribute.} The gap between easy and hard examples is not obvious, while the gap between confidence scores is large.
    }
    \vspace{-4mm}
    \label{easy hard sup 2}
\end{figure*}

Figure~\ref{easy hard sup 1} and Figure~\ref{easy hard sup 2} illustrate examples of images generated by providing easy and hard attributes. When $[hard]$ attributes are provided, the confidence score of the object is decreased to varying degrees, or some objects are missed in the detection. The changes in the objects in Figure~\ref{easy hard sup 1} are particularly noticeable. For instance, in (a), the confidence score is significantly reduced due to the reflection on the monitor screen. Additionally, changes in color, blurring of text, occlusion, and deformation lead to decreased confidence scores and missed detection in other examples.

The changes in the objects in Figure~\ref{easy hard sup 2} are minimal, yet they have a significant impact on the detector. This highlights the challenging examples that cannot be observed by the human eye but greatly affect the detector, which is what we aim to learn through attributes. For instance, in (a) and (c), only changes in color lead to a significant decrease in confidence score, while in (b), a minor change results in a substantial decrease in confidence score. Moreover, in (d) and (e), there are only slight deformations. Due to the sensitivity of detectors to subtle features, the use of prior-constructed image variations may not be effective for such cases. Our approach, which directly utilizes detection information to generate images, can reflect these differences, thereby further enhancing training.

\begin{figure*}[t]
	\centering
	\includegraphics[width=\linewidth]{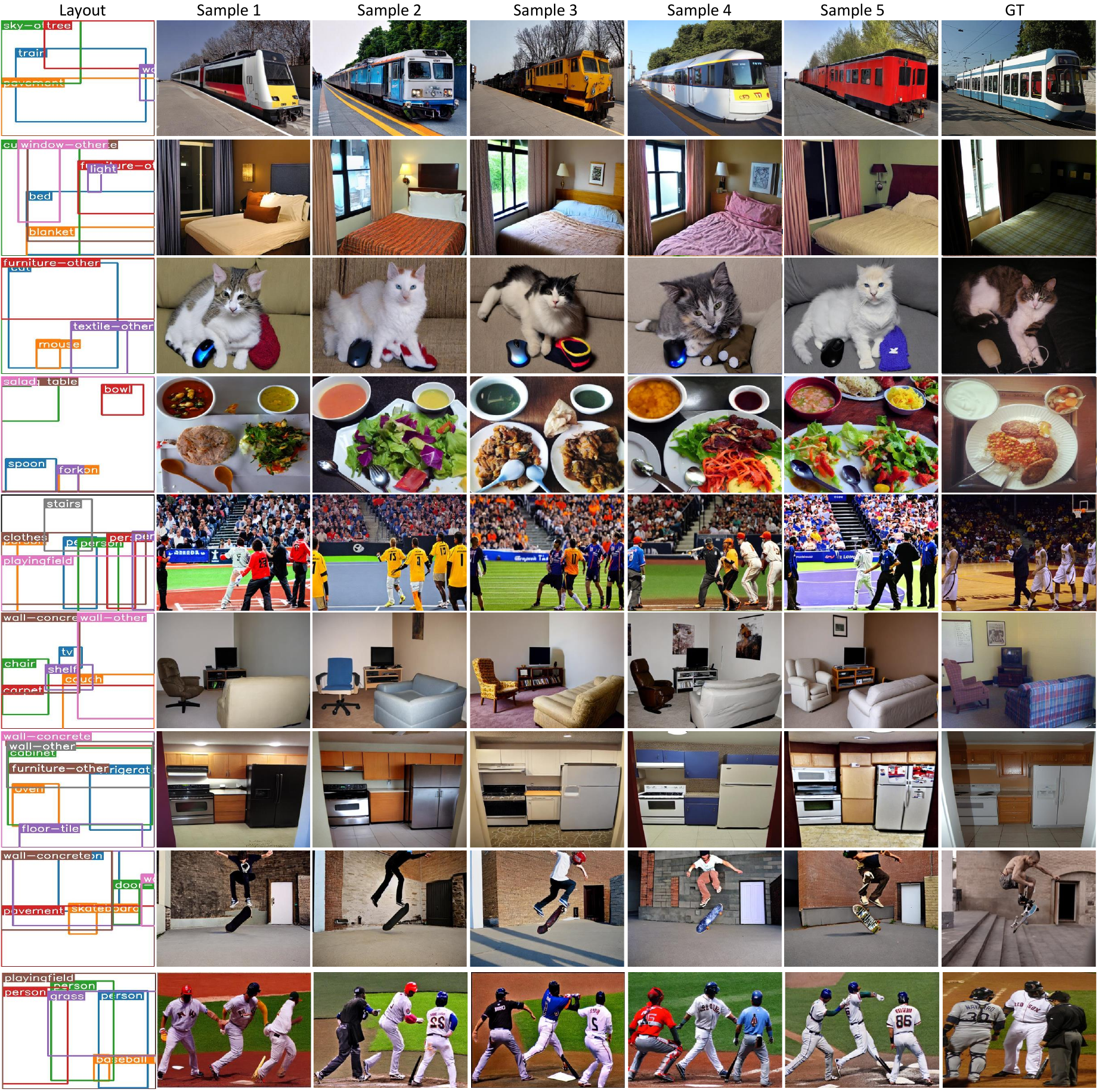}
    \caption{
    \textbf{More qualitative results of the same layout with random noises.} 
    }
    \label{diversity}
\end{figure*}

\methodname is capable of generating diverse scenes in Figure~\ref{diversity}, thus demonstrating our fidelity and diversity.



\end{document}